\documentclass{article}
\usepackage{ijcai25}

\usepackage{times}
\usepackage{soul}
\usepackage{url}
\usepackage[hidelinks]{hyperref}
\usepackage[utf8]{inputenc}
\usepackage[small]{caption}
\usepackage{amsmath}
\usepackage{amssymb} 
\usepackage{graphicx}
\usepackage{hyperref}  
\usepackage{cleveref}  
\usepackage{amsthm}
\usepackage{booktabs}
\usepackage{multirow}  
\usepackage{subfigure}    
\usepackage{subcaption} 
\usepackage{algorithm}
\usepackage{algorithmic}
\usepackage[switch]{lineno}

\title{
Spotlighting Partially Visible Cinematic Language for Video-to-Audio Generation via Self-distillation
}

\author{
Feizhen Huang\and
Yu Wu\and
Yutian Lin\thanks{Corresponding author.}\And
Bo Du$^*$\\
\affiliations
School of Computer Science, Wuhan University\\
\emails
\{feizhenhuang, wuyucs, yutian.lin, dubo\}@whu.edu.cn
}

\begin{document}

\maketitle

\begin{abstract}

Video-to-Audio (V2A) Generation achieves significant progress and plays a crucial role in film and video post-production. However, current methods overlook the \textit{cinematic language}, a critical component of artistic expression in filmmaking. As a result, their performance deteriorates in scenarios where Foley targets are only partially visible. To address this challenge, we propose a simple self-distillation approach to extend V2A models to cinematic language scenarios. By simulating the cinematic language variations, the student model learns to align the video features of training pairs with the same audio-visual correspondences, enabling it to effectively capture the associations between sounds and partial visual information. Our method not only achieves impressive improvements under partial visibility across all evaluation metrics, but also enhances performance on the large-scale V2A dataset, VGGSound.

\end{abstract}    
\section{Introduction}

\label{sec:intro}

Video-to-Audio (V2A) Generation~\cite{diff,perform1,V2Amapper,conFoley}, which generates corresponding audio directly from silent videos, has significant applications in film and video post-production.

\begin{figure}[!t]
    \centering
    \includegraphics[width=\linewidth]{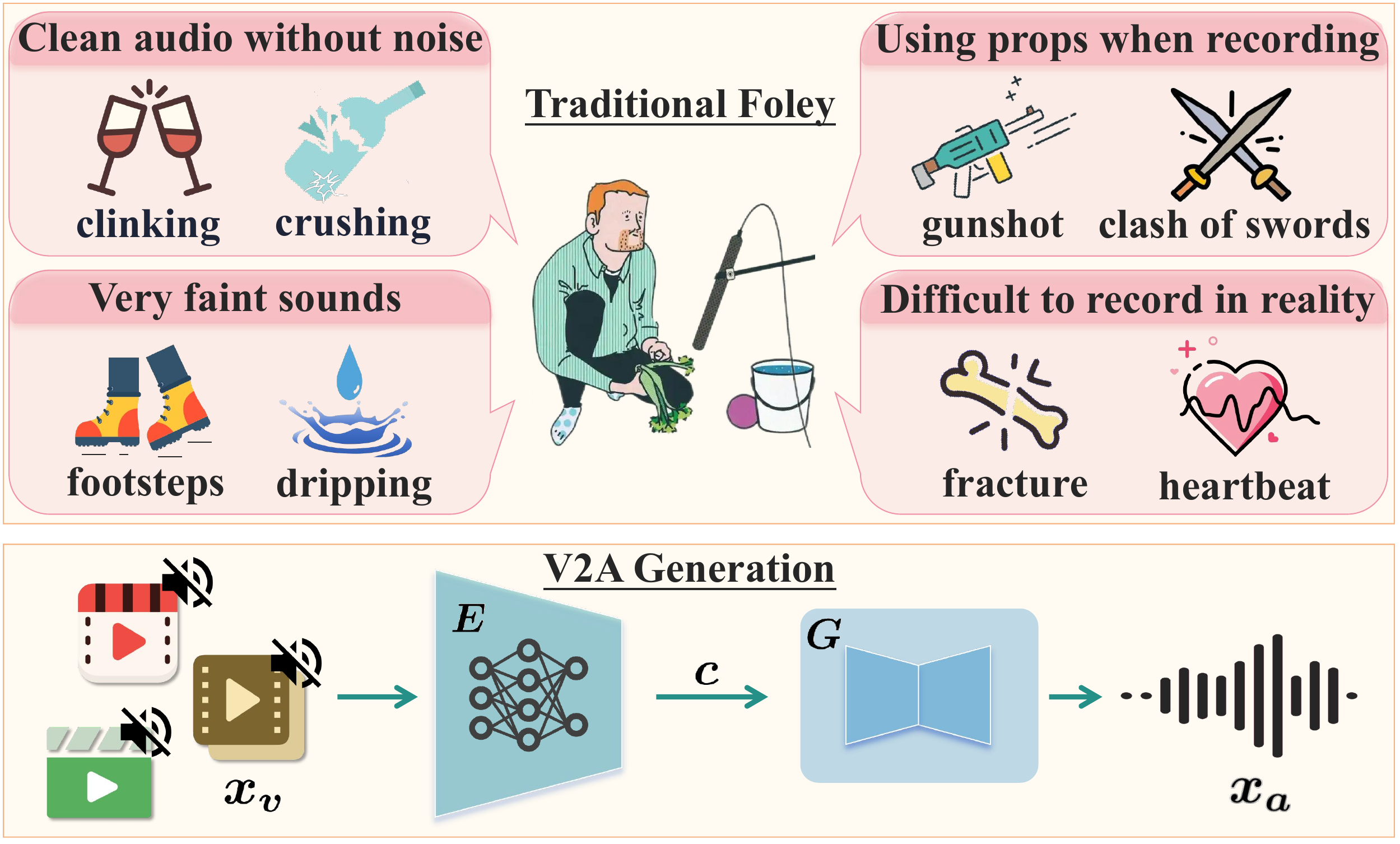}
    \caption{\textit{Foley} is the process of adding sound effects to silent videos, playing an essential role in film/video production due to factors illustrated in the pink boxes. 
    Traditional Foley relies on skilled Foley artists to manually reproduce sounds, whereas V2A Generation can directly generate corresponding audio from silent videos, providing a more efficient and convenient solution.
    }
    \label{fig:task}
\end{figure}

During live filming, capturing clean sound is often challenging due to ambient noise interference, the faintness of certain sounds, and other factors. As a result, most sounds must be recreated in post-production. As illustrated in \Cref{fig:task}, the process of adding relevant and synchronized sound effects to silent videos is known as \textit{Foley}~\cite{foley}. Traditional Foley requires skilled Foley artists to reproduce different sounds by manipulating various objects. The success of this process heavily depends on the artist's expertise and experience. Moreover, due to the vast array of sound categories and manipulable objects, this process is labor-intensive and time-consuming. This complexity hinders large-scale replication and individual video creation. In contrast, V2A Generation~\cite{diff,perform1,V2Amapper,conFoley} offers an appealing alternative by automatically generating corresponding audio directly from silent videos. This innovative approach alleviates the burden on human labor and offers a more scalable and faster solution for both individuals and companies.

Significant progress has been made in V2A Generation. SpecVQGAN~\cite{SpecVQGAN} leads the way in audio generation for open-domain videos. Diff-Foley~\cite{diff} stands out for its focus on addressing the challenge of audio-visual temporal synchronization, sparking a wave of subsequent research with impressive advancements in model performance~\cite{perform1,perform2,perform3}, precise temporal alignment~\cite{syn1,syn2}, lightweight designs~\cite{V2Amapper}, and text control integration~\cite{conFoley,control1,control2}. However, current methods overlook the role of \textit{cinematic language}~\cite{lan}, a cornerstone of artistic expression in film and video.

\begin{figure*}[!t]
    \centering
    \includegraphics[width=0.8\linewidth]{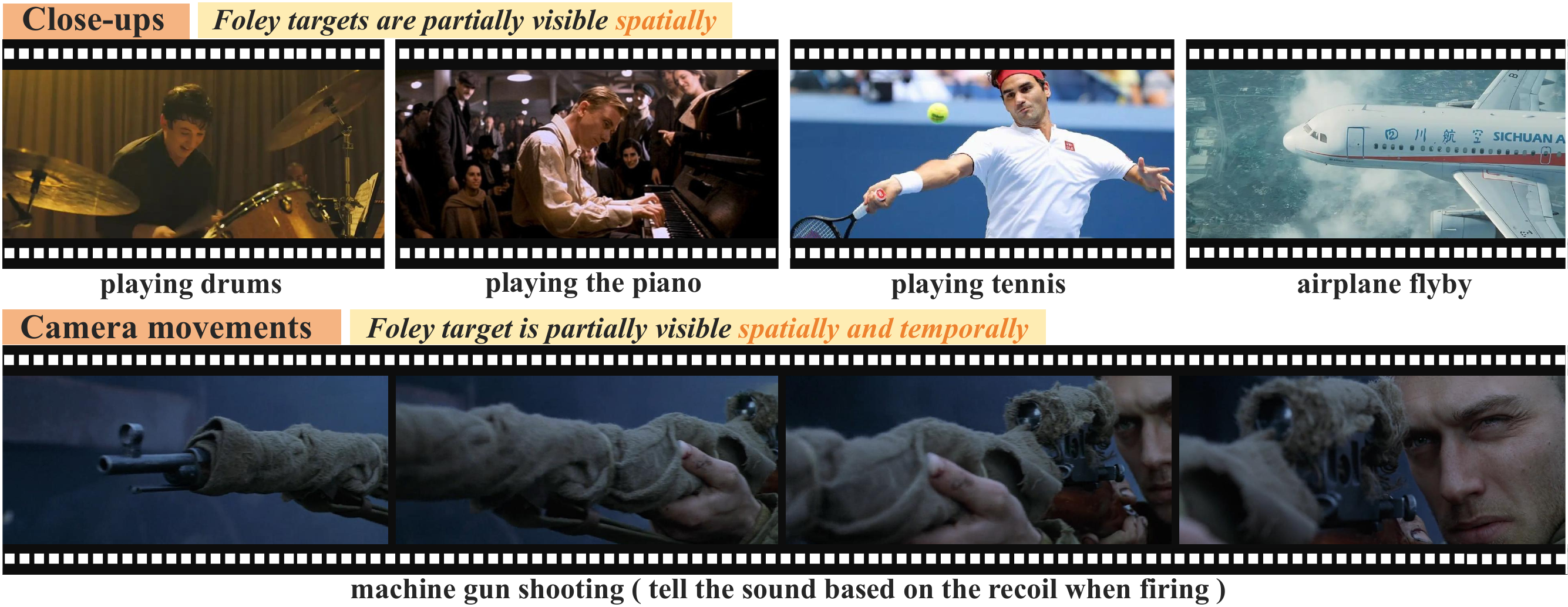}
    \caption{\textit{Cinematic language} is a fundamental element of artistic expression in film, such as close-ups and camera movements. These camera techniques often create scenarios where Foley targets are only partially visible spatially or temporally. We refer to them as \textbf{partial visibility}.}
    \label{fig:motivation}
\end{figure*}

\textit{Cinematic language}~\cite{lan} has the expressive power of storytelling, enabling directors to convey subjective intentions effectively. Camera techniques such as close-ups and camera movements are common to cinematic language. For instance, close-ups highlight specific features of characters or objects by zooming in, while camera movements dynamically introduce or remove them from the frame. These camera techniques aim to portray characters as shown in \Cref{fig:motivation}, such as characters playing instruments or engaging in sports, where Foley targets may only be \textit{partially visible in the spatial frame or temporal sequence}. We refer to these situations where Foley targets are partially visible spatially or temporally as \textbf{partial visibility}. In these situations, humans can infer sound based on other visual clues and temporal clues, even with incomplete information. However, current V2A methods struggle to handle the challenges posed by partial visibility, leading to poor performance in such scenarios.

State-of-the-art (SOTA) V2A models~\cite{diff,perform1,perform2} typically employ a two-stage training process: large-scale pretraining on audio-visual datasets to learn a robust video encoder, followed by training an audio generator conditioned on the extracted video features. In cinematic language scenarios, partial visibility of Foley targets causes the video encoder to extract inaccurate video features. This misrepresentation results in incorrect audio-visual associations, hindering the generation of corresponding audio. To directly adapt these models for cinematic scenarios, a straightforward approach would be collecting cinematic videos to retrain both the video encoder and the audio generator. 
However, this approach faces substantial obstacles. High-quality cinematic video clips are not only scarce but also restricted by copyright. Furthermore, training models directly on cinematic videos with incomplete visual information can be problematic. Such outliers may cause the model puzzled, rather than enabling it to effectively learn from partial visual information, ultimately disrupting optimization and resulting in unexpectedly poor performance.
Interestingly, current V2A models exhibit strong performance on non-cinematic videos. This suggests that the pre-trained knowledge can be leveraged to bridge the performance gap between non-cinematic and cinematic scenarios.

In this paper, we propose a simple teacher-student framework to capture partial audio-visual clues by constructing paired supervision video, focusing on addressing the challenges of partial visibility in cinematic scenarios. First, we simulate fundamental cinematic language variations to create paired training videos: one with cinematic variations and the other without while preserving consistent audio-visual correspondences. The latter serves as a supervision signal, subtly guiding the model to transfer its prior knowledge to understand the partial visual information. These paired videos are more effective than directly training on collected cinematic videos. Next, we adopt a teacher-student framework to align the video features from these paired training videos, benefiting from the supervision signals. This not only enables the student model to learn the associations between sounds and partial visual clues but also preserves its original performance.
Our approach is both efficient and general, requiring neither additional cinematic data nor modifications to the subsequent audio generative model.
It achieves impressive improvements under partial visibility across all evaluation metrics and enhances performance on the large-scale V2A dataset, VGGSound~\cite{diff} compared to the baseline. 
Our contributions are as follows:


\begin{itemize}
    \item We are the first to focus on cinematic language in the V2A generation area, where the Foley targets are only partially visible.
    
    \item We propose an efficient and general teacher-student framework that captures partial audio-visual clues by creating paired training videos while maintaining its original performance.
\end{itemize}
\section{Related Work}

\label{sec:related_work}

\subsection{V2A Generation}

The progress in V2A Generation research attracts a lot of attention. Early methods~\cite{RegNet} typically train separate models for each video category to generate more relevant and higher-fidelity audio, which limits the generalization abilities of models. SpecVQGAN~\cite{SpecVQGAN} makes a pioneering effort in audio generation for open-domain videos. Diff-Foley~\cite{diff} stands out by addressing the challenge of audio-visual temporal synchronization, inspiring a wave of subsequent studies with impressive results in enhanced model performance~\cite{perform1,perform2,perform3}, precise temporal alignment~\cite{syn1,syn2}, lightweight designs~\cite{V2Amapper}, and the integration of text control~\cite{conFoley,control1,control2}. Despite these advancements, current methods all neglect cinematic language and perform poorly in such scenarios, which is the focus of our work.

\subsection{Partial Visual Clues Perception}

Partial visual clues often arise when the predicted targets are partially occluded or out of sight. The ability to perceive partial visual information is a common challenge across various visual tasks. In traditional computer vision tasks, such as image classification and segmentation, models~\cite{image1,image2} need to make inferences based on visible portions. For person re-identification, models~\cite{reid1,reid2} are required to identify the same individual despite variations caused by camera perspectives and occlusions. Similarly, robots need to predict human motion directions and trajectories based on visible human body parts~\cite{robot1}. Scene understanding tasks~\cite{scene1,scene2} also face this challenge, as models must infer semantic information from incomplete visual clues, especially in applications like autonomous driving. In our research, we encounter similar difficulties related to partial visibility in V2A generation.

\begin{figure*}[!t]
    \centering
    \includegraphics[width=0.8\linewidth]{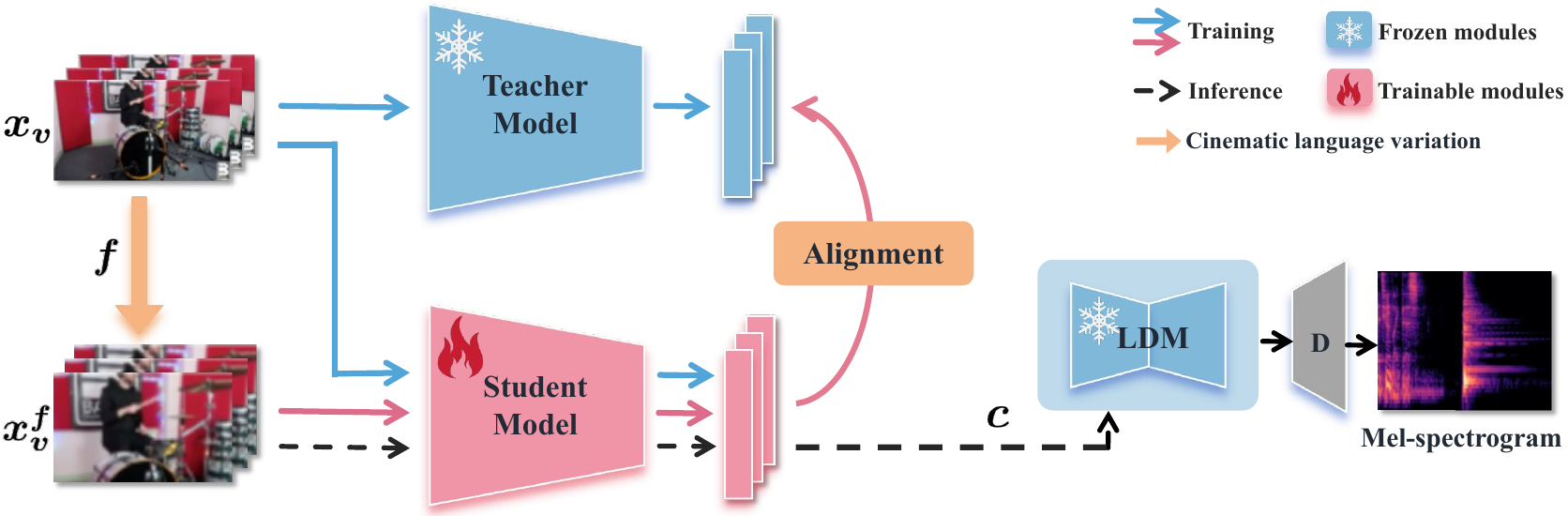}
    \caption{Our approach involves two key components: cinematic language variations $ f $ and a teacher-student framework. First, cinematic language variations $ f $ create training video pairs $ (x_v, x_v^f) $, where $ x_v^f $ retains the same semantic and temporal information of Foley target as $ x_v $. 
    By aligning the visual features extracted from $ x_v^f $ with those from $ x_v $ can facilitate the student model in learning the audio-visual connection with partial visibility. After training, the student model can provide a better generation condition for the subsequent inference stage
    }
    \label{fig:method}
\end{figure*}

\subsection{Teacher-Student Methodology}

The Teacher-Student methodology, also known as knowledge distillation~\cite{ts}, involves training a student model to learn from a teacher model. This approach aims to achieve model compression or enhance performance. Self-distillation~\cite{self1,self2,self3,self4} is a unique variant of the Teacher-Student methodology, where the teacher and student models are different versions or stages of the same model. This characteristic enables the model to learn from itself. In our work, we adopt this practical training methodology, leveraging the prior knowledge of a pre-trained video encoder to guide the learning of audio-visual correlation under partial visibility.

\subsection{Data Augmentation}

Data augmentation is widely applied to improve model performance and generalization abilities. In image-related fields, data augmentation techniques are widely used in traditional computer vision tasks~\cite{image1,image2}, such as flipping, rotation, translation, and noise injection. Moreover, video-based data augmentation methods ~\cite{video1} also gain attention in video action recognition. These methods concentrate more on actions and temporal information by altering color~\cite{vcolor} or blending foregrounds and backgrounds~\cite{vforeback,vforeback2,vforeback3} from different videos. Our method simulates cinematic language variations to emphasize audio-visual alignment using partial visual clues.

\section{Method}

An overview of our proposed method is shown in ~\Cref{fig:method}. First, we introduce the motivation behind our method in ~\Cref{sec:motivation}. Next, we propose cinematic language variations $ f $ in ~\Cref{sec:clvar} and adopt the teacher-student methodology to learn partial visibility in ~\Cref{sec:pvts}. Finally, in ~\Cref{sec:ldms}, we describe the latent diffusion models for V2A Generation with cinematic language.

\subsection{Motivation}
\label{sec:motivation}

Our method targets the challenge of partial visibility in cinematic language scenarios. To do so, we aim to improve the model's ability to capture audio-visual corrections from partial visual information.

As illustrated in \Cref{fig:task}, most Video-to-Audio (V2A) models follow a two-stage process: first, a feature extractor $ E $ extracts visual features $ c = E(x_v) $ from the video $ x_v $. Subsequently, using these visual features $ c $ as a condition, a generative model $ G $ generates the corresponding audio $ x_a = G(c) $. Indeed, the performance of the generative model $ G $ relies heavily on the semantic and temporal information embedded in the condition $ c $. Therefore, the quality of the condition $ c $ is critical for generating relevant and synchronized audio.

For normal video clips $ x_v $, i.e., \textit{without} cinematic language, the generative model $ G $ performs well, indicating that the extracted visual features $ c=E(x_v) $ provide accurate audio information, making them ideal generation conditions. However, when $ x_v' $ \textit{with} cinematic language, the quality of the generated audio $ x_a' $ deteriorates significantly. Given the strong performance of $ G $ on non-cinematic videos, we reasonably infer that this degradation stems from the poor generation conditions $ c' = E(x_v') $, rather than the generative model $ G $ itself. In other words, when the Foley targets are only partially visible, the video encoder $ E $ struggles to capture meaningful audio-visual associations from partial visual clues, thereby failing to provide a suitable generation prior.

Consequently, assisting the video encoder in learning the associations between sounds and partial visual clues and providing better conditions for the generative model $ G $ emerges as a natural approach.

\subsection{Cinematic Language Variations}
\label{sec:clvar}

We propose Cinematic Language Variations $ f $ to simulate partial visibility in cinematic language~\cite{lan}, helping to establish audio-visual associations in such scenarios. This is more effective than directly training on collected cinematic videos. In cinematic language, close-ups and camera movements are two common camera techniques that cause situations where the Foley targets are partially visible in the spatial or temporal dimension. The cinematic language variations $ f $ simulates the process of these two techniques as examples.

Given that close-ups emphasize specific local details of characters or objects, $ f_{cu} $ uniformly crops the video frames. The cropping sizes $ H' $ and $ W' $ are randomly selected within a reasonable range, which are determined as follows:
\begin{align}
    H' = H \times r_h, \quad W' = W \times r_w,
\end{align}
where $ H $ and $ W $ are the original height and width of the video frames, and $ r_h, r_w \sim U(a_1, a_2) $. The range $ [a_1, a_2] $ should be chosen to ensure that the Foley targets are partially visible while retaining sufficient visual information.

To simulate camera movements, $ f_{cm} $ utilizes the same cropping size above and shifts the shot along the central axis of the video frames. The shift direction is:

\begin{itemize}
    \item When $ W > H $, shift left or right at random.
    \item When $ H > W $, shift up or down at random.
\end{itemize}


\subsection{Partial Visibility Learning by Self-distillation}
\label{sec:pvts}

We adopt the teacher-student methodology to learn the partial visibility simulated by Cinematic Language Variations $ f $. Given that the pre-trained video encoder contains rich audio-visual priors, we utilize it as the teacher model $ T $ to guide the student model $ S $ in learning the association between sounds and partial visual clues, as illustrated in ~\Cref{fig:method}.

Given video data $ x_v $, by simulating cinematic language variations $ f $, we obtain video $ x_v^f $. Derived from $ x_v $, $ x_v^f $ retains the same semantic and temporal information of the Foley target with $ x_v $. They constitute training video pairs $ (x_v, x_v^f) $ with the same audio-visual correspondence. The only difference between training video pairs is the partial visibility of the Foley target: it is fully visible in video $ x_v $ while partially visible in video $ x_v^f $. Since the Foley target in video $ x_v $ is fully visible, the visual features $ c_t = T(x_v) $ extracted by the teacher model $ T $ accurately encompass sufficient audio information for conditioning. Thus, aligning the features $ c_{s^f} = S(x_v^f) $ extracted by the student model $ S $ with feature $ c_t $ can facilitate the student model $ S $ in learning the audio-visual connection between sounds and partial visual clues. During training, we also align $ c_s=S(x_v) $ with the feature $ c_t $ to maintain the performance on the original dataset. The optimization loss $ L_p $ is defined as follows:
\begin{align}
    L_p = \cos(c_t, c_{s'}) + \text{MSE}(c_t, c_{s'}), \quad c_{s'} = c_s \text{ or } c_{s^f}
\end{align}

We introduce $ k $ as the proportion of training video clips with cinematic language variations. By aligning the visual features of training video pairs, we can effectively guide the student model $ S $ to map videos with partial visibility into the original feature space of the teacher model. This approach not only learns a better generation condition but also requires no modifications to the subsequent generative model.

\begin{table*}[!t]
    \centering
    \small
    \begin{tabular}{ll|rrcrrc}
        \toprule
        Test Set & Method & FAD$\downarrow$ & FD$\downarrow$ & KID$(10^{-3})\downarrow$ & KL$\downarrow$ & IS$\uparrow$ & Align Acc(\%)$\uparrow$ \\
        \midrule
        \multirow{2}{*}{VGGSound} & Diff-Foley & 7.481 & 25.445 & 10.71 & 3.280 & 11.670 & \textbf{92.946} \\
        & Ours & \textbf{7.173} & \textbf{24.532} & \textbf{10.05} & \textbf{3.235} & \textbf{11.835} & 91.718 \\
        \midrule
        \multirow{2}{*}{VGG-CU} & Diff-Foley & 8.926 & 28.843 & 12.17 & 3.824 & 11.136 & 74.882 \\
        & Ours & \textbf{7.825} & \textbf{25.661} & \textbf{10.45} & \textbf{3.438} & \textbf{11.601} & \textbf{85.071} \\   
        \midrule
        \multirow{2}{*}{VGG-CM} & Diff-Foley & 9.164 & 28.394 & 11.48 & 3.843 & 10.665 & 72.570 \\
        & Ours & \textbf{8.194} & \textbf{25.932} & \textbf{10.23} & \textbf{3.508} & \textbf{11.047} & \textbf{81.850} \\
        \bottomrule
    \end{tabular}
    \caption{Evaluation results for Video-to-Audio generation across three test sets: the original VGGSound test set, VGG-CU (close-up) test set, and VGG-CM (camera movement) test set. During training, only cinematic language variation $ f_{cu} $ is applied to VGGSound~\protect\cite{vggsound} training set with $ k=75\% $. Diff-Foley baseline exhibits a notable performance drop on both VGG-CU and VGG-CM test sets, compared to its original performance on VGGSound test set. In contrast, our method significantly outperforms the baseline across all evaluation metrics on these two test sets, suggesting that it learns the audio-visual associations under partial visibility.}
    \label{tab:results_v2a}
\end{table*}

\subsection{Latent Diffusion Models}
\label{sec:ldms}

Latent Diffusion Models (LDMs)~\cite{LDM} are probabilistic generative models that map the data distribution into a low-dimensional latent space, consisting of an auto-encoder and a U-Net denoiser. In the V2A task, the latent encoder $ E $ encodes the Mel-spectrogram $ x \sim p(x) $ into the latent representation $ z = E(x) $, and the UNet denoiser $ \epsilon_\theta $ is then trained to reverse the noise addition to generate new latent representations. Under a given condition $ c $, the optimization~\cite{DDPM,DDIM} process of LDMs can be defined as follows:
\begin{align}
    L_{LDM} := \mathbb{E}_{z \sim \mathcal{E}(z), c, \epsilon \sim \mathcal{N}(0,1), t} \left[ \left\| \epsilon - \epsilon_\theta(z_t, t, c) \right\|_2^2 \right],
\end{align}
where $ \epsilon $ represents Gaussian noise, and $ z_t $ denotes the latent representation at time step $ t $.

For conditional LDMs, classifier-free guidance (CFG) ~\cite{CFG} is a widely used alternative to classifier guidance (CG)~\cite{CG}. CFG jointly trains conditional $ \epsilon_\theta(x_t, c, t) $ and unconditional $ \epsilon_\theta(x_t, t) $ diffusion models by randomly dropping the condition $c$. During sampling, the noise prediction is calculated as:
\begin{align}
    \hat{\epsilon}_\theta(z_t, c, t) = w \epsilon_\theta(z_t, c, t) + (1 - w) \epsilon_\theta(z_t, t),
\end{align}
where $w$ is the guidance scale. 

In our study, we build upon the open-source Diff-Foley~\cite{diff}, an LDM conditioned on video features extracted from a frozen CAVP video encoder, to achieve Foley in cinematic scenarios.

\section{Experiments}

\paragraph{Datasets.} We conduct our experiments using VGGSound ~\cite{vggsound}, a large-scale audio-visual dataset containing over 200,000 video clips across 309 distinct sound categories. Existing V2A methods utilize VGGSound for both training and evaluation. We follow the original VGGSound train/test split.

\paragraph{Baseline.} Given our choice of Diff-Foley~\cite{diff} as the foundation model for our study, we select Diff-Foley as our baseline, which is a leading open-source V2A model. 
For simplicity, we adopt the CFG configuration for both generation and comparison, keeping all other settings unchanged.

\paragraph{Evaluation Metrics.} For evaluation, we adopt evaluation metrics FAD, FD, KID, KL, and ISc from audioLDM~\cite{audioldm}, alongside Align Acc from Diff-Foley~\cite{diff}. FAD, FD, and KID measure the similarity between real and generated audio, while KL assesses the paired similarity in probability distributions. ISc measures the diversity and quality of generated audio, and Align Acc assesses the audio-visual synchronization.

\begin{figure}[!t]
    \centering
    \subfigure[Close-ups]{
        \includegraphics[width=0.47\textwidth]{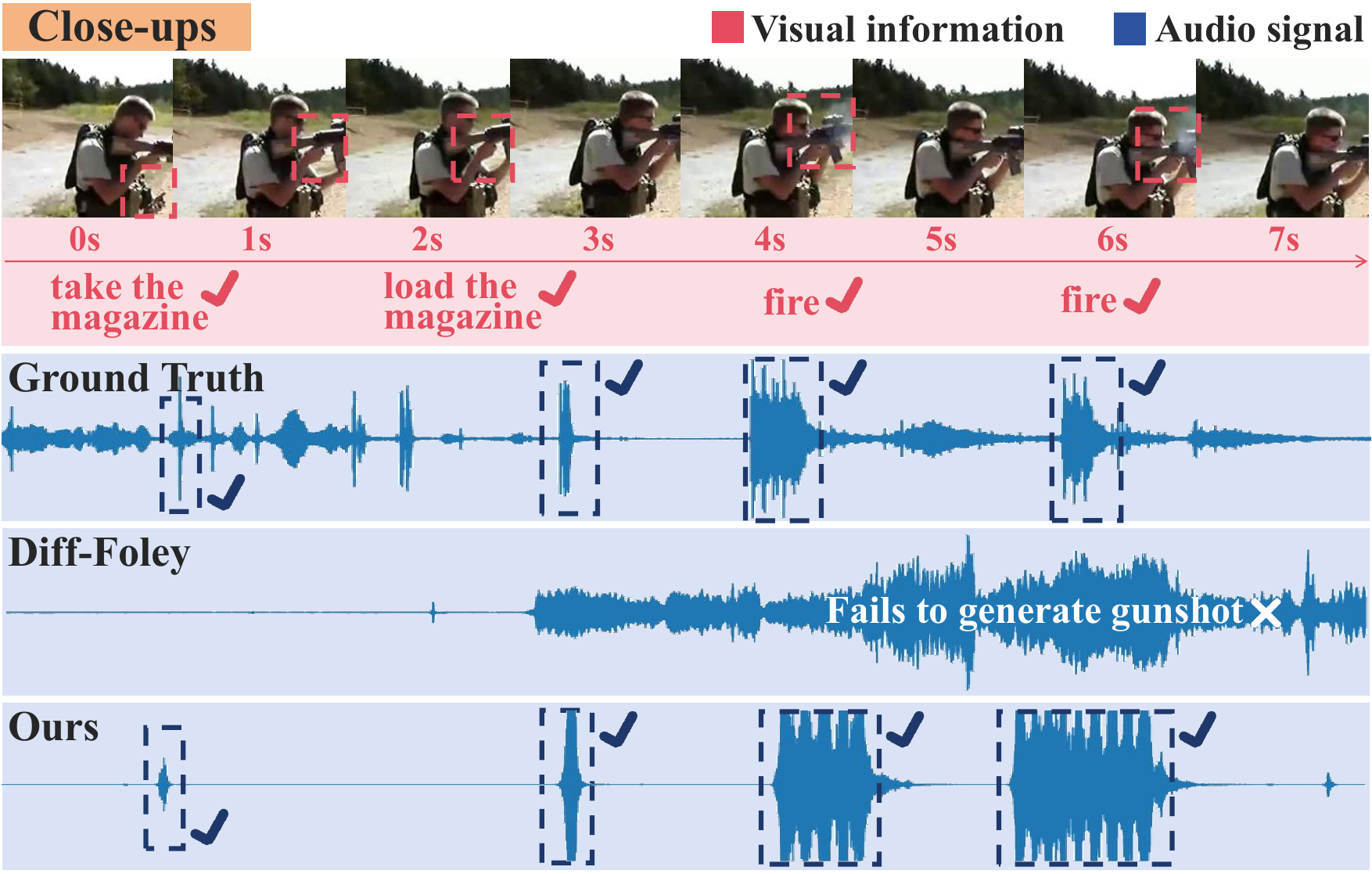}
        \label{fig:v2a_cu}
    }
    \hfill
    \subfigure[Camera Movements]{
        \includegraphics[width=0.47\textwidth]{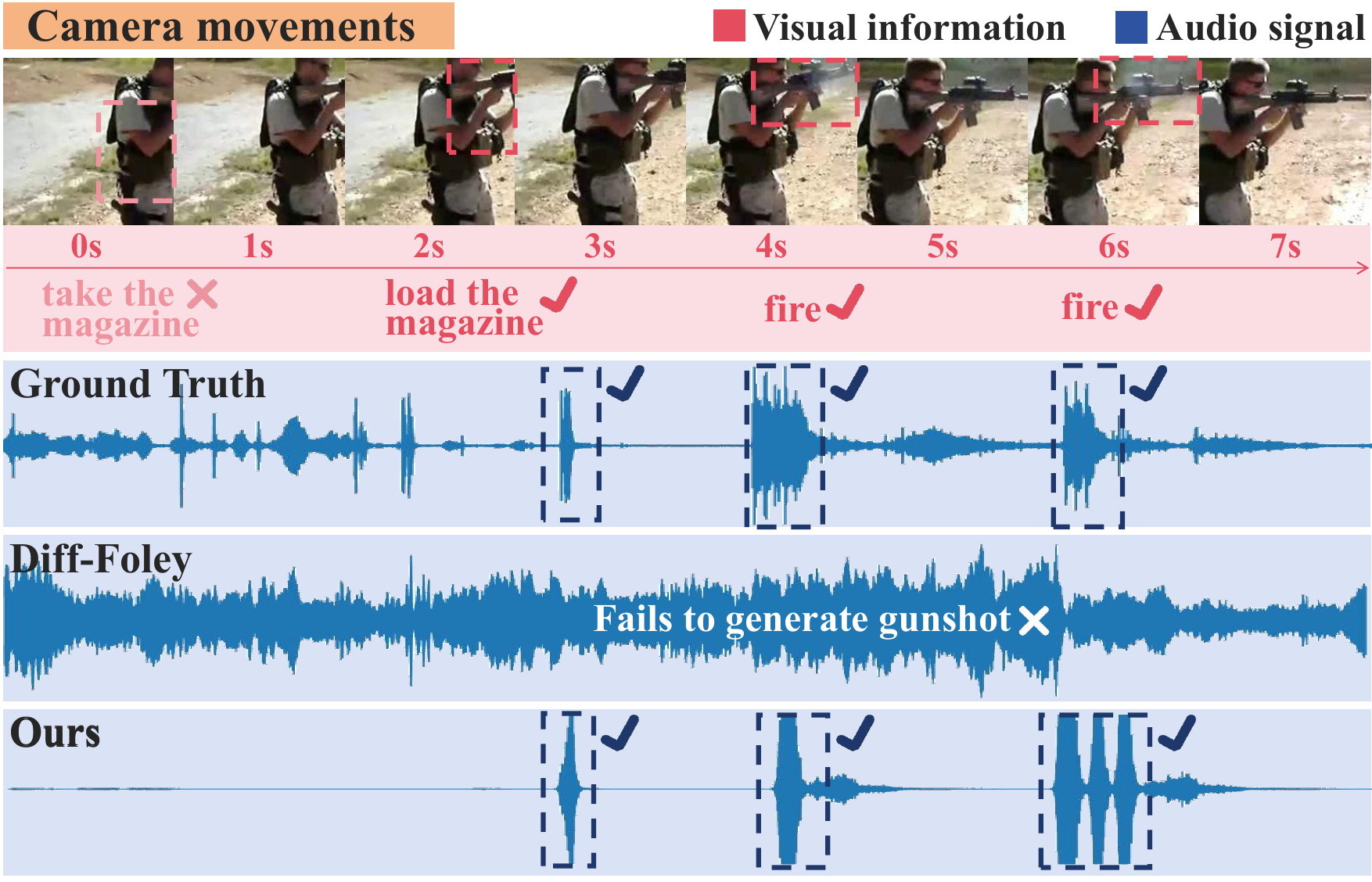}
        \label{fig:v2a_cm}
    }
    \caption{The figures show the qualitative results for V2A generation in cinematic language scenarios involving close-ups and camera movements. Taking a machine gun shooting video as an example, the pink dashed boxes and text mark the partial visual information, while the blue parts represent the corresponding auditory signals.}
    \label{fig:v2a_cucm}
\end{figure}

\begin{figure*}[!t]
    \centering
    \includegraphics[width=0.85\linewidth]{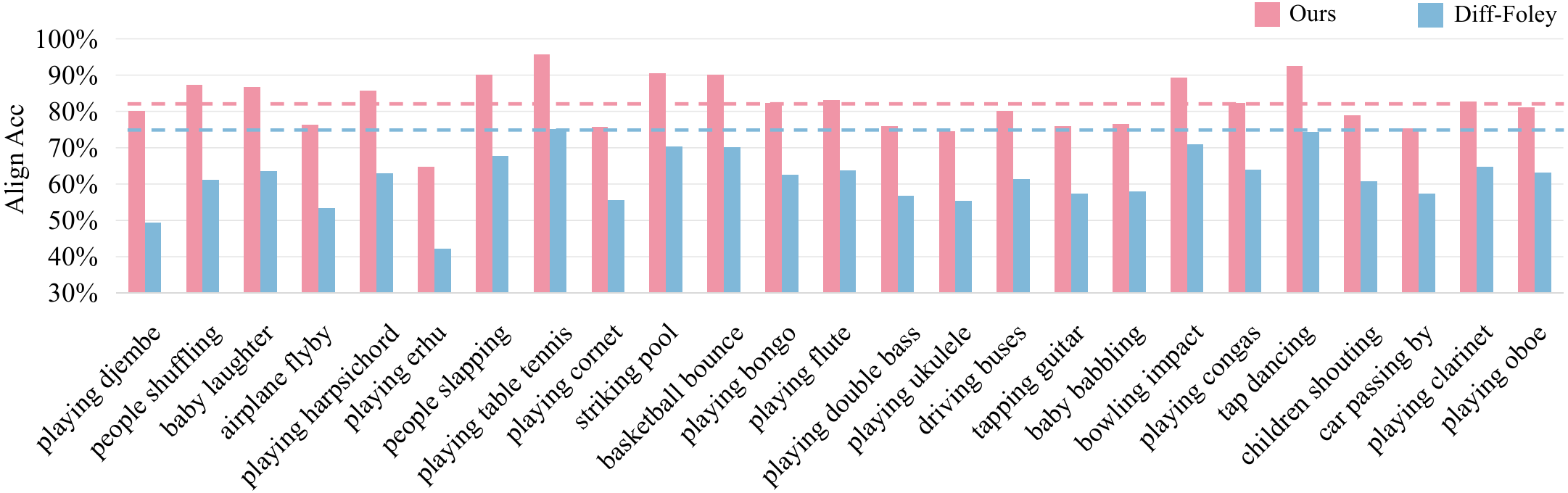}
    \caption{We evaluate the Align Acc metric for each video category on VGG-CU test set and present the top 25 video categories (309 in total) with the most significant improvements. The dashed line represents the overall Align Acc across the entire VGG-CU test set.}
    \label{fig:topcls}
\end{figure*}

\paragraph{Implementation Details.} For model configuration, we employ a pre-trained video encoder from CAVP~\cite{diff} as the teacher model. The student model adopts the same architecture, with weights initialized from the teacher model’s pre-trained parameters. The input video clips are sampled at 4 frames per second (FPS), resulting in $ T=4N $ frames for each $N$-second video clip. Then the input video $ x_v \in \mathbb{R}^{T \times 3 \times H \times W }$ is extracted by the video encoder into video feature $ E_v \in \mathbb{R}^{T \times C}$, with the feature dimension $ C=512 $.
For training, we only apply cinematic language variation $ f_{cu} $ on VGGSound~\cite{vggsound} training set with $ k=75\% $, where $ a_1=0.4 $ and $ a_2=0.6 $. The student model is trained for 25 epochs on 4 NVIDIA 4090 GPUs, using the AdamW optimizer with a learning rate of $5 \times 10^{-4}$ and a total batch size of 32.  

To evaluate performance under partial visibility, we create two modified test sets by applying cinematic language variations to the VGGSound~\cite{vggsound} test set. Specifically, $ f_{cu} $ is used to create VGG-CU (close-ups) test set, and $ f_{cm} $ is used to create VGG-CM (camera movements) test set. Following Diff-Foley~\cite{diff}, we generate 10 samples per video to ensure reliable evaluation. For simplicity, we use only CFG~\cite{CFG} configuration in Diff-Foley, keeping all other experimental settings unchanged, including the DPM-Solver~\cite{dpm} Sampler with 25 inference steps and CFG scale $ \omega = 4.5 $.

\subsection{V2A Generation with Cinematic Language}
\label{sec:results_v2a}


\paragraph{Simulated Cinematic Scene on VGGSound.} \Cref{tab:results_v2a} presents the quantitative results across three test sets: the original VGGSound test set, VGG-CU test set (created using $f_{cu}$), and VGG-CM test set (created using $f_{cm}$). In cinematic language scenarios, we observe that Diff-Foley~\cite{diff} baseline shows a significant decline in all evaluation metrics on both VGG-CU and VGG-CM test sets compared to its performance on the original VGGSound test set. 
In contrast, our method outperforms the baseline by a substantial margin across all evaluation metrics on these two test sets. Notably, even when trained only on VGG-CU training set, our model exhibits considerable improvement on VGG-CM test set. 
These results suggest that our method effectively learns the associations between audio and partial visual information. Furthermore, by learning partial visual information, our method achieves improved performance on the original VGGSound test set compared to the baseline.

\begin{table}[!t]
    \centering
    \begin{tabular}{cc}
        \toprule
        prefer Diff-Foley & prefer Ours \\
        \midrule
        158 (25.48\%) & 462 (74.51\%) \\   
        \bottomrule
    \end{tabular}
    \caption{A human study is conducted using collected 31 YouTube videos with diverse cinematic scenarios from real-world.}
    \label{table:human}
\end{table}

~\Cref{fig:v2a_cucm} presents the qualitative results for V2A generation in cinematic language scenarios involving close-ups and camera movements. Taking the machine gun shooting video as an example, partial visual information is highlighted with pink dashed boxes and pink text, while the corresponding auditory signals are represented in blue. As depicted in~\Cref{fig:v2a_cucm}, we can easily identify the machine gun and the moment of the sound from the video frames, even when the gun is partially visible. However, Diff-Foley fails to generate the appropriate gunshot and other sounds at the correct moment. In contrast, our method successfully generates synchronized gunshot audio when the Foley target is partially visible. Notably, in camera movements scenarios shown in ~\Cref{fig:v2a_cm}, the visual clues in the 0s and 1s video frames are lost. Our method does not generate sound when the visual clues are completely missing. While in the following video frames, our method correctly generates relevant and synchronized audio. These results suggest that our method correctly learns the audio-visual associations under partial visibility.


\paragraph{Real Cinematic Scene.} To evaluate performance in real-world cinematic scenarios, we collect a new set of 31 YouTube videos that feature a wide range of cinematic techniques, ensuring no overlap with the original VGGSound test set. These videos include characteristics such as close-ups, camera movements, zooms, scene transitions, as well as non-partial factors like color and lighting variations. We conducted a human study with 20 participants and the results confirm the superiority of our method as shown in \Cref{table:human},

\subsection{Analysis of Partial Visibility Learning}

\begin{figure}[!t]
    \centering
    \includegraphics[width=\linewidth]{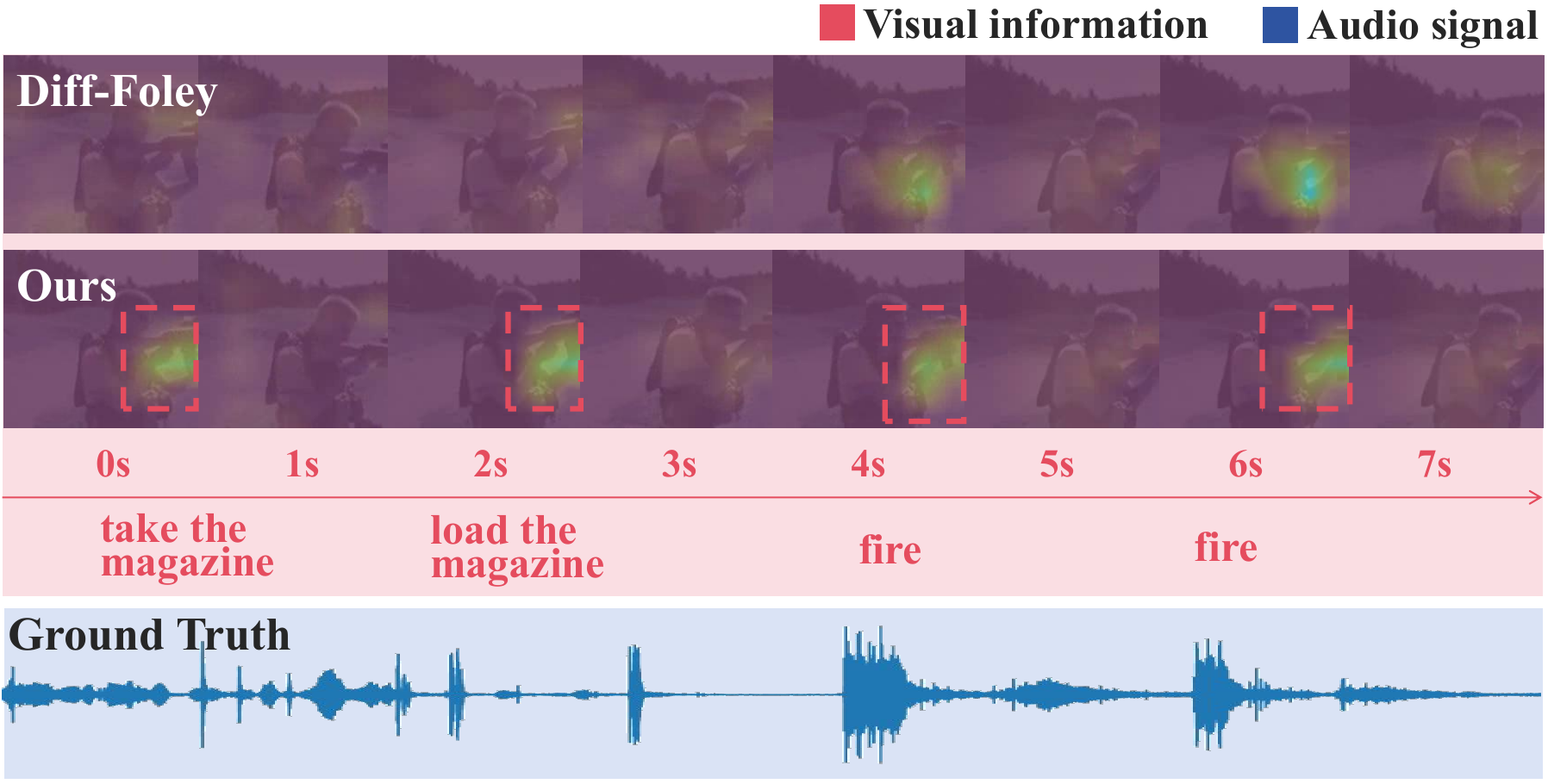}
    \caption{We use the Grad-CAM~\protect\cite{gradcam} to visualize the model’s attention in close-up scenarios. Taking the machine gun shooting video as an example, the pink dashed boxes and text represent the partial visual information, while the corresponding audio ground truth is represented in blue.}

    \label{fig:heatmap}
\end{figure}

\paragraph{Visualization Analysis.} To better understand what the model learns, we visualize the model's attention areas along the video frames in close-up scenarios. As shown in \Cref{fig:heatmap}, we observe that the four regions highlighted by the model's attention (denoted by pink dashed lines) are semantically relevant to the partial visual clues. Compared to the ground truth audio, these highlighted regions are also temporally aligned with the corresponding sound. In contrast, the highlighted regions of Diff-Foley's attention are misaligned with the partial visual clues and are not temporally synchronized with the audio. These visualized results demonstrate that our model accurately captures the semantic and temporal association between sound and partial visual clues.


\begin{table*}[!t]
    \centering
    \begin{tabular}{ll|rrcrrc}
        \toprule
        Test Set & Training $f$ & FAD$\downarrow$ & FD$\downarrow$ & KID$(10^{-3})\downarrow$ & KL$\downarrow$ & IS$\uparrow$ & Align Acc(\%)$\uparrow$ \\
        \midrule
        \multirow{3}{*}{VGG-CU} & $f_{cu}$ & \textbf{7.905} & \textbf{26.341} & \textbf{10.65} & \textbf{3.511} & 11.411 & 82.951 \\
        & $f_{cm}$ & 8.505 & 27.403 & 11.30 & 3.652 & 11.645 & 79.420 \\
        & $f_{cu} \& f_{cm}$ & 8.157 & 26.460 & 10.76 & 3.513 & \textbf{11.705} & \textbf{83.311} \\
        \midrule
        \multirow{3}{*}{VGG-CM} & $f_{cu}$ & \textbf{8.311} & \textbf{26.861} & \textbf{10.59} & 3.585 & 10.918 & 79.437 \\
        & $f_{cm}$ & 8.699 & 27.523 & 11.19 & 3.624 & \textbf{11.418} & 79.115 \\
        & $f_{cu} \& f_{cm}$ & 8.653 & 27.057 & 11.02 & \textbf{3.558} & 11.311 & \textbf{80.514} \\
        \midrule
        \multirow{3}{*}{VGGSound} & $f_{cu}$ & \textbf{7.174} & \textbf{24.686} & \textbf{10.17} & \textbf{3.242} & 11.739 & 91.398 \\
        & $f_{cm}$ & 7.306 & 24.917 & 10.38 & 3.257 & \textbf{11.933} & \textbf{91.560} \\
        & $f_{cu} \& f_{cm}$ & 7.388 & 24.889 & 10.33 & 3.261 & 11.895 & 91.119 \\
        \bottomrule
    \end{tabular}
    \caption{We explore the effect of different cinematic language variations $ f $ during training, including $f_{cu}$ (close-ups), $f_{cm}$ (camera movements), and $f_{cu\&cm}$ (a combination of both). The respective proportions are set as $ k_{f_{cu}}=50\% $, $ k_{f_{cm}}=50\% $, and $k_{f_{cu\&cm}}=66.7\%$ (with $f_{cu}:f_{cm}=1:1$). The evaluation settings remain consistent with the previous experiments.}
    \label{tab:results_cmcn}
\end{table*}

\begin{table*}[!t]
    \centering
    \begin{tabular}{ll|rrcrrc}
        \toprule
        Test Set & Proportion $k$ & FAD$\downarrow$ & FD$\downarrow$ & KID$(10^{-3})\downarrow$ & KL$\downarrow$ & IS$\uparrow$ & Align Acc(\%)$\uparrow$ \\
        \midrule
        \multirow{3}{*}{VGG-CU} & 50\% & 7.905 & 26.341 & 10.65 & 3.511 & 11.411 & 82.951 \\
        & 75\% & \textbf{7.825} & \textbf{25.661} & \textbf{10.45} & \textbf{3.438} & 11.601 & \textbf{85.071} \\
        & 100\%  & 7.992 & 26.135 & 10.67 & 3.468 & \textbf{11.727} & 84.121 \\
        \midrule
        \multirow{3}{*}{VGG-CM} & 50\% & 8.311 & 26.861 & 10.59 & 3.585 & 10.918 & 79.437 \\
        & 75\% & \textbf{8.194} & \textbf{25.932} & \textbf{10.23} & \textbf{3.508} & 11.047 & \textbf{81.850} \\
        & 100\% & 8.393 & 26.366 & 10.46 & 3.520 & \textbf{11.130} & 81.279 \\
        \midrule
        \multirow{3}{*}{VGGSound} & 50\% & 7.174 & 24.686 & 10.17 & 3.242 & 11.739 & 91.398 \\
        & 75\% & \textbf{7.173} & \textbf{24.532} & \textbf{10.05} & \textbf{3.235} & \textbf{11.835} & \textbf{91.718} \\
        & 100\% & 8.036 & 26.487 & 10.68 & 3.425 & 11.454 & 86.553 \\
        \bottomrule
    \end{tabular}
    \caption{We explore the proportion $ k $ of training video clips with cinematic language variation $f_{cu}$.}
    \label{tab:results_ratio}
\end{table*}

\paragraph{Improvements in Different Video Categories.} To further understand where our approach is superior, We evaluate the Align Acc metric for each video category on VGG-CU test set and present the top 25 video categories (309 in total) with the most significant improvements. As shown in \Cref{fig:topcls}, these categories primarily include instrumental sounds, ball impact sounds, and vehicle movement sounds that are commonly found in cinematic scenarios. In these categories, Diff-Foley exhibits notably lower audio-visual consistency compared to its average performance. In contrast, our method achieves significantly better results, demonstrating its effectiveness in addressing the challenges of partial visibility in cinematic scenarios.

\subsection{Ablation Study}


\paragraph{The Impact of Different Variations.} As shown in \Cref{tab:results_cmcn}, we introduce three cinematic language variations during training to assess their impact: $f_{cu}$, $f_{cm}$, and $f_{cu\&cm}$. The $f_{cu}$ variation simulates spatial partial visibility, while $f_{cm}$ simulates temporal partial visibility. According to FAD, FD, and KID metrics, only using $f_{cu}$ yields the best improvements across all three test scenarios, suggesting that the generated audio is semantically more aligned with the video. This indicates that simulating spatial partial visibility ($f_{cu}$) is more effective than simulating temporal partial visibility ($f_{cm}$) for learning the semantic association between sounds and partial visual clues. On the other hand, in terms of Align ACC, $f_{cu\&cm}$ achieves the best improvements in VGG-CU and VGG-CM test sets, with the generated audio showing better temporal alignment with the videos. This suggests that simulating both spatial and temporal partial visibility ($f_{cu\&cm}$) enhances the model's ability to learn the temporal synchronization between sounds and partial visual clues.

\paragraph{The Proportion of Variation.} \Cref{tab:results_ratio} demonstrates the effect of the proportion $ k $ of training video clips with cinematic language variation $f_{cu}$. In VGG-CU and VGG-CM test sets, model performance improves as $ k $ increases from 50\% to 75\%, showing the effectiveness of $f_{cu}$. When $ k $ reaches 100\%, where no data from the original dataset is used, the model's performance shows a decline compared to 75\%, emphasizing the importance of training with the paired data.

\section{Limitations and Broader Impact}

\paragraph{Limitations.}
Our method excels in scenarios with partial visibility such as camera movements and close-ups, but its adaptability and performance across other diverse cinematic languages still require further exploration. 

\paragraph{Broader Impact.}
V2A boosts video production efficiency, offering substantial benefits to creators. However, vigilance and regulations are still needed to prevent potential misuse.

\section{Conclusion}

We present a simple yet effective method to address the challenge of partial visibility in cinematic language scenarios by providing a better condition for the audio generator. By simulating the cinematic language variations, the student model aligns the video features from training video pairs with the same audio-visual correspondences, which helps learn the associations between sounds and partial visual clues. Experimental results demonstrate the impressive improvements of our approach, not only in partial visibility scenarios but also on the original V2A dataset, VGGSound.

\section*{Acknowledgments}
This work was partially supported by the Key Research and Development Program of Yunnan Province under Grant 202403AA080002, partially supported by the National Natural Science Foundation of China under Grant 62471344, and partially supported by the National Natural Science Foundation of China under Grant 62372341.

\bibliographystyle{named}
\bibliography{ijcai25}

\end{document}